\documentclass[letterpaper, 10 pt, journal, twoside]{IEEEtran}   
\usepackage{hyperref}
\hypersetup{colorlinks = true, allcolors = cyan}
\usepackage{color,xcolor}
\usepackage{graphicx,times,amsmath,amssymb,mathrsfs,subfigure,stfloats,booktabs,url,multirow}
\usepackage[lined,ruled,commentsnumbered,linesnumbered]{algorithm2e}
\definecolor{ccr}{RGB}{10,110,150}  
\usepackage[T1]{fontenc}
\usepackage{graphicx,times,amsmath,amssymb,mathrsfs,subfigure,stfloats,booktabs,url,multirow}
\usepackage{color,xcolor}
\begin{document}
\title{Co-Fusion4D: Spatio-temporal Collaborative Fusion for Robust 3D Object Detection}
\author{Wenxuan Li, Qin Zou\, Shoubing Chen, Chi Chen, Yingyi Yang,  Qingxiang Meng}
\maketitle              
\begin{abstract}
In autonomous driving, 3D object detection is essential for accurate perception and reliable decision-making. However, object motion and ego-motion often induce cross-frame spatiotemporal inconsistencies in BEV-based detectors, leading to temporal BEV feature misalignment and degraded spatiotemporal consistency.
To address these challenges, we propose Co-Fusion4D, a unified framework that explicitly preserves cross-frame spatiotemporal consistency and suppresses temporal feature drift. Co-Fusion4D adopts a current-frame–centric strategy, treating the current frame as the primary source of information while selectively incorporating historical frames after spatiotemporal filtering and alignment. This dominant–complementary mechanism effectively mitigates cumulative alignment errors, suppresses noisy feature propagation, and exploits reliable temporal cues for a more consistent BEV representation.
In addition, Co-Fusion4D integrates a Dual Attention Fusion (DAF) module to further enhance spatiotemporal feature interaction. DAF jointly leverages intra-frame spatial attention and inter-frame temporal attention to adaptively align and fuse multi-frame features, emphasizing motion-consistent regions while suppressing spurious correlations. By departing from conventional uniform fusion paradigms, this design substantially improves the temporal stability and discriminative capability of BEV representations.
Extensive experiments on the nuScenes benchmark demonstrate that Co-Fusion4D achieves state-of-the-art performance, with 74.9\% mAP and 75.6\% NDS, without relying on test-time augmentation or external data.

\begin{IEEEkeywords}
	3D	object detection, multimodal fusion, temporal BEV, autonomous driving, deep learning
\end{IEEEkeywords}
\end{abstract}
\section{Introduction}

Accurate and robust environmental perception is fundamental for autonomous driving, where 3D object detection~\cite{song2024robustness,mu2025stereodetr} plays a pivotal role in understanding the surrounding dynamic environment and enabling reliable decision-making. Multi-modal fusion, which integrates the precise geometric measurements from LiDAR with rich semantic cues from cameras into Bird’s Eye View (BEV) representations~\cite{li2025bevfix,feng2025iter3ddet}, has emerged as an effective strategy for enhancing spatial perception. Such fusion not only facilitates accurate object localization and classification but also supports downstream tasks such as multi-object tracking and motion prediction~\cite{ma2024vision,jia2025dgfusion}.

Recent advances\cite{zhou2025fastpillars} have focused on \textit{temporal BEV-based 3D object detection}, aiming to exploit multi-frame information to improve motion awareness and temporal consistency. However, maintaining spatiotemporal alignment across frames remains a significant challenge, especially in multi-modal scenarios. Temporal inconsistencies arise due to ego-vehicle motion, dynamic objects, and varying sensor viewpoints. Misaligned or outdated BEV features from historical frames can propagate errors, degrading detection performance. Moreover, sensor noise, motion blur, discretization artifacts, and environmental variations further compromise the reliability of temporal feature representations.

Existing temporal BEV approaches partially address these challenges but remain limited. For instance, BEVFusion4D~\cite{cai2023bevfusion4d} employs sequential chain-based attention to capture temporal dependencies but suffers from substantial computational overhead, limiting real-time applicability. GAFusion~\cite{li2024gafusion} performs simple concatenation of multi-frame BEV features without explicit motion compensation, leading to feature redundancy and semantic interference in dynamic scenes. Image-only temporal BEV detectors such as BEVFormer, StreamPETR, and UniBEV leverage global cross-frame attention or recurrent BEV memory, yet they are prone to feature drift, temporal instability, and high computational cost. Collectively, these limitations highlight the need for methods that can explicitly ensure spatiotemporal consistency, robustly handle dynamic motion, and balance accuracy with efficiency.

To address these challenges, we propose \textbf{Co-Fusion4D}, a temporal multi-frame 3D detection framework. Inspired by Co-Fix3D~\cite{li2024cofix3d}, Co-Fusion4D enhances BEV features through multiple parallel Local-Global Enhancement (LGE) modules, while introducing a \textit{current-frame-centric strategy} that prioritizes the most reliable information from the current frame and selectively incorporates historical frames. We further introduce a \textbf{Dual Attention Fusion (DAF)} module, which jointly leverages intra-frame spatial attention and inter-frame temporal attention. Intra-frame static attention captures long-range spatial dependencies, improving localization of objects in static regions. Inter-frame dynamic attention models temporal evolution, enabling robust detection and tracking of moving objects. This dual mechanism ensures effective temporal feature fusion and enhances the discriminative power of multi-frame BEV representations.

Overall, Co-Fusion4D is an end-to-end temporal BEV-based framework that combines multi-modal fusion with spatiotemporally consistent multi-frame aggregation. A four-stage training strategy is employed to mitigate overfitting and stabilize optimization across LiDAR, camera, and temporal domains. Extensive experiments on the nuScenes dataset demonstrate that Co-Fusion4D achieves state-of-the-art performance, attaining \textbf{74.9\% mAP} and \textbf{75.6\% NDS} without relying on test-time augmentation or external data.

The main contributions of this work are summarized as follows:
\begin{itemize}
	\item We propose \textbf{Co-Fusion4D}, a temporal multi-frame 3D object detection framework that integrates sequential LiDAR point clouds and camera images for robust dynamic scene perception.
	\item We design a \textbf{four-stage training strategy} that decouples modality-specific learning and temporal feature optimization, enhancing stability and overall detection performance.
	\item We introduce a Dual Attention Fusion (DAF) module that jointly models intra-frame spatial and inter-frame temporal dependencies, improving feature consistency and dynamic object detection.
	\item Extensive experiments on the nuScenes benchmark validate the effectiveness of Co-Fusion4D, demonstrating superior performance over existing state-of-the-art multi-modal 3D detectors.
\end{itemize}

\section{Related Work}

\subsection{LiDAR-based 3D Detectors}

LiDAR-based 3D object detection has been extensively studied due to its accurate geometric perception capability in autonomous driving. Existing methods can be broadly categorized into point-based, voxel-based, and hybrid approaches. Point-based methods such as PointNet~\cite{qi2017pointnet} and PointNet++~\cite{qi2017pointnet++} directly process raw point clouds to learn fine-grained geometric representations, and are further extended by proposal-based frameworks like PointRCNN~\cite{shi2019pointrcnn}. Voxel-based methods, including VoxelNet~\cite{zhou2018voxelnet} and SECOND~\cite{yan2018second}, discretize point clouds into structured grids to enable efficient convolutional operations. Hybrid methods such as PV-RCNN~\cite{shi2020pv} combine the advantages of point-wise precision and voxel-wise efficiency, achieving strong performance on large-scale benchmarks.

Despite their effectiveness, LiDAR-only detectors primarily rely on geometric cues and often struggle in challenging scenarios such as long-range detection, occlusion, and sparse point distributions. Moreover, they typically operate on single-frame inputs and lack temporal modeling capability, limiting their robustness in dynamic environments. These limitations motivate the integration of complementary modalities and temporal information.

\subsection{Multi-Modal 3D Detectors}

Multi-modal 3D object detection\cite{chen2024dsc3d,li2025tinyfusiondet} aims to combine LiDAR and camera data to leverage complementary information, where LiDAR provides accurate depth and structure, while cameras contribute dense semantic and texture cues. Existing fusion strategies are generally categorized into early, middle, and late fusion. Early fusion methods directly combine raw sensor data but are sensitive to calibration errors~\cite{chen2022deformable,xu2021fusionpainting}. Middle fusion methods integrate features at intermediate layers, achieving a good balance between performance and robustness~\cite{li2022voxel,li2022unifying}. Late fusion combines high-level predictions but often limits deep cross-modal interaction.

Recent advances focus on BEV-based multi-modal fusion, where features from different modalities are unified in the Bird’s Eye View space. Representative methods such as TransFusion~\cite{bai2022transfusion} and BEVFusion~\cite{liang2022bevfusion,liu2023bevfusion} utilize attention mechanisms to align and fuse multi-modal features, achieving state-of-the-art performance. Recent surveys highlight that BEV-centric fusion and transformer-based architectures have become dominant trends in multi-modal 3D detection~\cite{wang2023multi,singh2023transformer}. 
However, most existing multi-modal approaches are designed for single-frame perception and do not fully exploit temporal information across frames. As a result, they may fail to capture motion cues and temporal consistency, which are critical for dynamic object detection. Co-Fusion4D addresses this limitation by extending multi-modal fusion into the temporal domain through consistent multi-frame BEV representation learning.

\subsection{Temporal and Multi-Frame Fusion for 3D Detection}

Temporal modeling has emerged as a key direction for improving 3D object detection in dynamic driving environments. By leveraging historical frames, temporal methods can capture motion patterns, alleviate occlusion, and enhance detection robustness. 

In image-based BEV detection, BEVFormer~\cite{li2022bevformer} introduces spatiotemporal transformers to aggregate multi-frame features, demonstrating strong performance in multi-view perception tasks. StreamPETR~\cite{wang2023exploring} further proposes an object-centric temporal modeling mechanism that propagates queries across frames, enabling efficient long-term temporal reasoning with reduced computational overhead.

In multi-modal settings, BEVFusion4D~\cite{cai2023bevfusion4d} extends BEV fusion into the temporal domain by aggregating historical BEV features with cross-attention mechanisms. More recent methods such as FocalFusion~\cite{wan2025focalfusion} explore object-centric temporal fusion strategies and motion-aware feature alignment to improve dynamic scene understanding. Despite these advances, many existing approaches rely on global fusion across frames, which can introduce feature misalignment, temporal noise accumulation, and high computational cost.

In contrast, Co-Fusion4D adopts a \textbf{current-frame-centric} temporal fusion paradigm, which treats the current frame as the primary reference and selectively incorporates aligned historical information. Furthermore, we propose a \textbf{Dual Attention Fusion (DAF)} module that decouples intra-frame spatial modeling from inter-frame temporal reasoning, enabling more stable and discriminative feature aggregation. Compared with prior methods, Co-Fusion4D effectively mitigates temporal feature drift, suppresses noisy propagation, and achieves a better balance between accuracy and efficiency for multi-modal 3D object detection in dynamic environments.

\section{Method}
\begin{figure*}[t]
	\centering
	\includegraphics[width=0.98\linewidth]{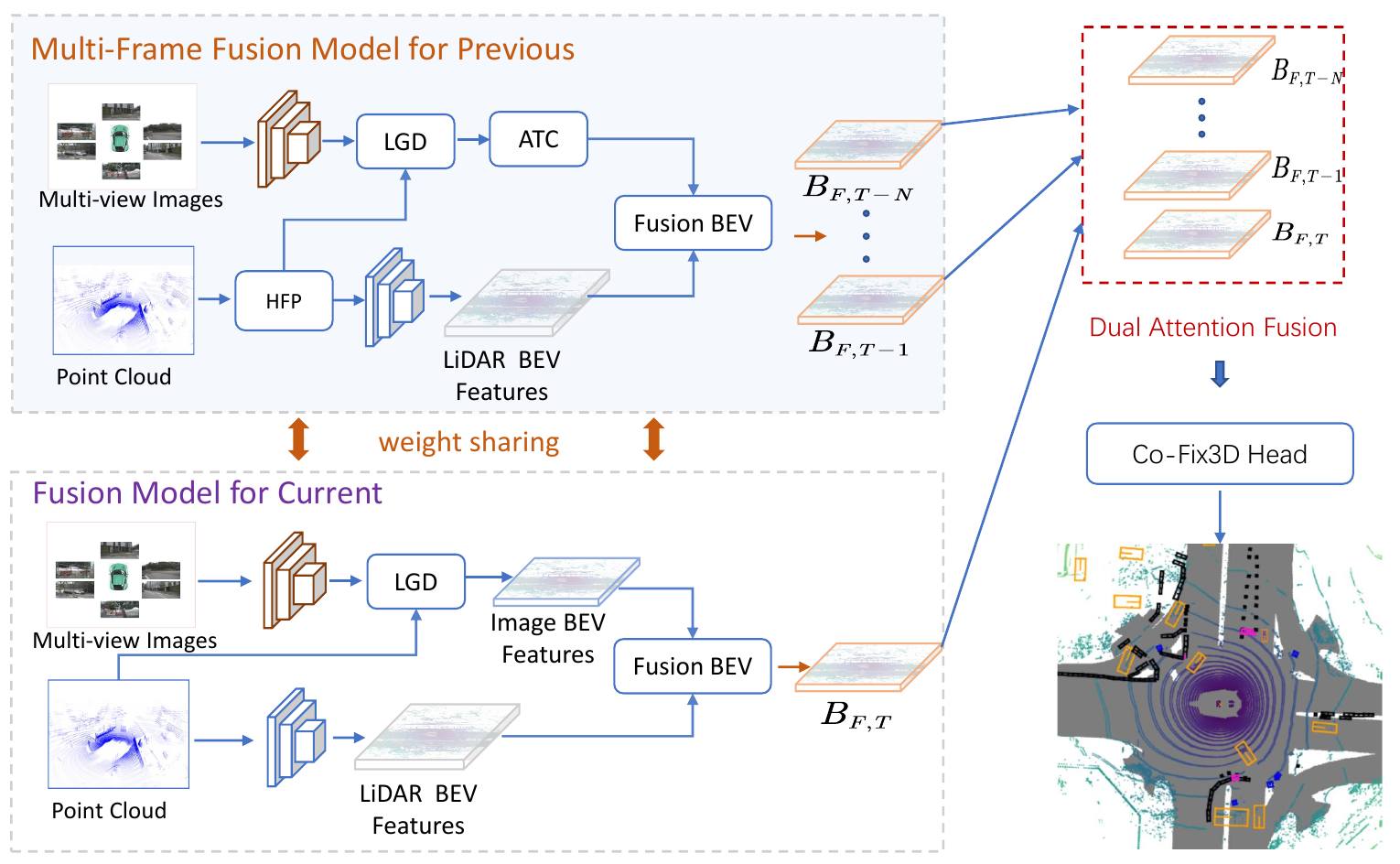}	
	\caption{							
	Overview of Co-Fusion4D: Co-Fusion4D integrates multi-frame, multimodal data, combining images and point clouds. For the current frame, it uses a dual-branch approach: the point cloud branch generates LiDAR BEV features, while the image branch processes data through the LiDAR-Depth Guidance (LDG) module to generate image BEV features. These features are then fused into a fused BEV representation. For historical frames, point clouds are aligned with the current frame using the Point-to-Current (PTC) module, and historical image data is aligned using the Align-to-Current (ATC) module, generating corresponding fused BEV features for each timestamp. These fused BEV features are enhanced via a multi-stage Local-Global Enhancement (LGE) module, with temporal fusion handled by the Dual Attention Fusion (DAF) module, which integrates temporal and spatial features across frames. Finally, predictions are made using the Co-Fix3D head: the Top-k method selects the highest-scoring BEV grids as queries, and a Transformer model decodes these queries to generate detection predictions.}
	\label{main}
	\vspace*{-4mm}
\end{figure*}

\subsection{Overall Framework}	

As illustrated in Fig.~\ref{main}, Co-Fusion4D adopts a dual-branch paradigm to effectively process cross-modality data, treating the current and historical frames differently. For the current frame, Co-Fusion4D follows the pipeline of Co-Fix3D: point clouds captured at time T and multi-view images from the same timestamp are processed. The LiDAR stream processes the point cloud, denoted as $P_T$. The point clouds are first voxelized and then processed using 3D sparse convolutions to generate a BEV representation, $B_{\text{L},T} \in \mathbb{R}^{X \times Y \times C_P}$, where $X$, $Y$, and $C_P$ represent the dimensions of the BEV grid and feature channels. This process can be expressed as follows:
\begin{equation}
	B_{\text{L}, T} = f_{\text{LiDAR}}(P_T),
	\label{eq:overall}
\end{equation}
where $f_{\text{LiDAR}}(\cdot)$ denotes the process of converting point clouds to BEV.

Concurrently, the image stream processes multi-view images to extract 2D features, $I_T \in \mathbb{R}^{N \times C_I \times H \times W}$, where $N$, $ C_I$, $H$, and $W$ represent the number of cameras, feature dimensions, image height, and width, respectively. And  the LiDAR-Depth Guidance (LDG) module, inspired by DAL~\cite{huang2024detecting}, fuses depth map features and image characteristics through convolution, and then converts these 2D features into BEV representations, aligning them with the LiDAR features.
\begin{equation}
  B_{\text{I,T}}=f_{\text{Camera}}(I_T),
\end{equation}
where $f_{\text{Camera}}(\cdot)$ denotes the process of converting multi-view images  to image BEV features. Then, the BEV features from the point cloud branch and the image branch are fused to form a new, unified multi-modal BEV feature representation. This process is denoted as:
\begin{equation}
	B_{\text{F,T}}=f_{\text{Fusion}}(B_{\text{L}, T},B_{\text{I,T}} )
\end{equation}
where $f_{\text{Fusion}}(\cdot)$ denotes the process of fusing the BEV features from the LiDAR branch and the image branch.

For historical frames, our model follows a similar process to that of the current frame. However, the point cloud and image data from historical frames must be projected into the current frame through coordinate transformations. Specifically, the PTC module is used to perform the coordinate transformation for the point cloud, while the ATC module is applied to transform the image data in the image branch. Then, a network with shared weights, used in conjunction with the current frame, generates the BEV features for historical timestamps, such as $B_{F,T-1}$, $B_{F,T-N}$. To fully exploit detection capabilities, particularly its ability to mine hard samples, Co-Fix3D employs a multi-stage approach with a mask mechanism to progressively filter each stage, enabling parallel LGE modules to supervise different ground truths. Each LGE module at each stage forms temporal BEV features according to the following equation:
\begin{equation}
	B_{i,T},B_{i,T-1},\cdot \cdot \cdot, B_{i,T-N}=f_{LGE_i}(B_{F,T},B_{F,T-1},\cdot \cdot \cdot, B_{F,T-N}),
\end{equation}
where $i$ represents the stage number in the multi-stage detection process, and each stage is associated with its corresponding LGE  module.
For each stage, we use the DAF module to fuse the temporal BEV features, resulting in the final BEV representation:

\begin{figure*}[h]
	
	\centering
	\includegraphics[width=0.7\linewidth]{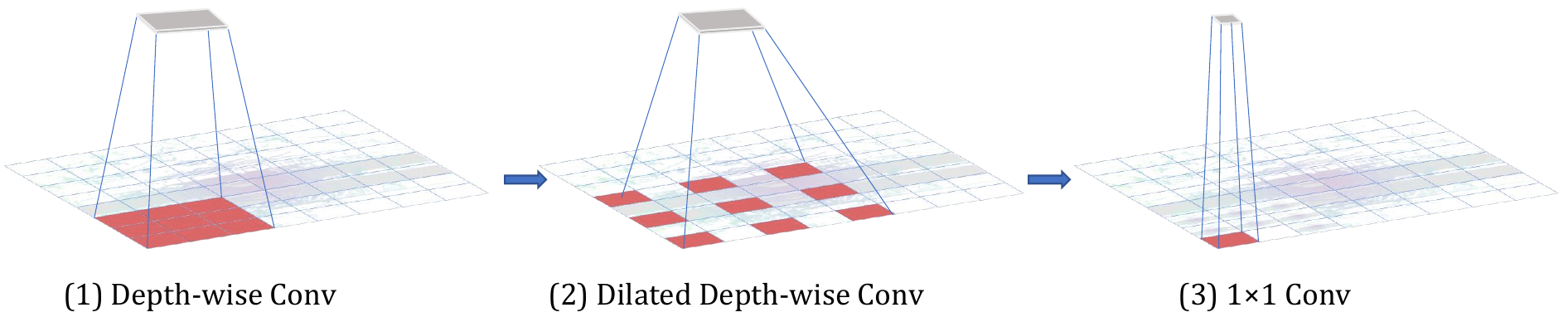}	
	\caption{Intra-frame Static Attention.						
	}
	\label{inter-f}
	\vspace*{-4mm}
\end{figure*}

\begin{equation}
	B_{i}=f_{DAF}(B_{F,T},B_{F,T-1},\cdot \cdot \cdot, B_{F,T-N}),
\end{equation}

The subsequent process follows the Co-Fix3D architecture, involving decoding and predicting object classes and locations. At each stage $i$, it applies Top-k selection to set $k$ instances of $M_i(w, h, c)$ to 0, ensuring that once a region is selected, subsequent stages won't re-explore it. Box-level pooling handles the 0-marked masks, ensuring an even distribution of query locations. If earlier modules fail to detect certain samples, later modules will continue to refine them. Through these mechanisms, queries are collected, decoded by the decoder, and passed through the prediction module for final predictions. The equations are as follows:

\begin{equation}
	R_{class}, R_{box}=f_{pred}({B_{1},B_{2},\cdot \cdot \cdot, B_{i}}),
\end{equation}
where $ f_{\text{pred}}(\cdot) $ denotes the process of making final predictions.

\subsection{Point Cloud Alignment and BEV Feature Alignment}

In multi-frame 3D object detection, geometric alignment across temporal steps is a fundamental prerequisite for stable temporal modeling. Due to the continuous motion of autonomous platforms and sensor systems, point clouds and images captured at different timestamps are defined in their respective local coordinate systems. Directly fusing historical data with current-frame features would introduce systematic spatial offsets, which break temporal consistency and significantly degrade the effectiveness of multi-frame information. Therefore, accurate spatio-temporal alignment is essential, both for point-level fusion and BEV-based feature-level temporal modeling.

\paragraph{Point Cloud Alignment (HFP).}
The goal of point cloud alignment is to map the 3D structure of historical frames into the coordinate system of the current frame, enabling point clouds from different timestamps to share a unified spatial reference. This process relies on the rigid transformation matrix 
$T^{k \rightarrow t} \in SE(3)$, obtained from extrinsic calibration and ego-motion estimation, which encodes both rotation and translation.

For any point $\mathbf{p}_k$ in the historical frame $k$, its position in the current frame $t$ is given by:
\begin{equation}
	\mathbf{p}_t = T^{k \rightarrow t} \, \mathbf{p}_k.
	\label{eq:point_align}
\end{equation}

Through this transformation, static structures (e.g., roads, buildings, and static obstacles) observed at different timestamps can be spatially aligned in 3D space, achieving cross-frame geometric consistency. This not only ensures the geometric accuracy of multi-frame point cloud fusion, but also provides stable input for subsequent BEV generation and global feature reasoning. 

Such alignment is particularly critical in scenarios involving high-speed motion, sharp turns, or vibrations, where even small pose estimation errors can be significantly amplified in spatial coordinates, adversely affecting temporal modeling quality.

\paragraph{BEV Feature Alignment (ATC).}
Beyond geometric alignment at the point level, multi-frame BEV-based perception also requires feature-level temporal consistency in the network space. The core idea of BEV feature alignment is to utilize the same transformation matrix $T^{k \rightarrow t}$ to map historical BEV features into the current BEV coordinate system.

Specifically, the spatial coordinates (or grid indices) of historical BEV features are transformed according to $T^{k \rightarrow t}$, enabling direct fusion with current-frame BEV representations. Unlike point cloud alignment, this operation is performed in the feature space after geometric projection, serving as a \emph{post-projection alignment} mechanism.

The necessity of BEV feature alignment arises from discretization effects introduced by voxelization and sparse convolution. Although point cloud alignment removes global coordinate discrepancies, these discretization processes inevitably introduce grid-level quantization errors. As a result, even geometrically aligned points may still exhibit sub-grid misalignment in BEV feature maps.

To address this issue, BEV feature alignment performs interpolation-based resampling in continuous coordinate space, effectively mitigating discretization-induced residual errors and improving temporal feature consistency.

\subsection{Dual Attention Fusion }
For the temporal BEV features, we use the Dual Attention Fusion (DAF) module for fusion, which consists of two main components: Intra-frame Static Attention and Inter-frame Dynamic Attention. We  now describe each component in detail:
	
\textbf{Intra-frame static attention} is designed to capture long-range dependencies within a single frame. To achieve this, we employ small kernel depth-wise convolutions (DW Conv), depth-wise convolutions with dilations (DW-D Conv), and 1×1 convolutions:

\begin{itemize}
	
	\item \textbf{DW Conv:} Focuses on applying convolution to individual channels, reducing computational overhead compared to traditional convolutions with large kernels.
	\item \textbf{DW-D Conv:} Extends this by incorporating dilations, which expands the receptive field without increasing the number of parameters, helping the model capture larger context and dependencies across the frame.
	\item \textbf{1×1 convolutions:} Serve as a channel aggregation mechanism, enabling the network to combine information from different feature maps, enhancing the depth of feature extraction.
\end{itemize}

These operations simulate the effect of large kernel convolutions by allowing the network to model both local and global relationships while maintaining computational efficiency.
The specific formula for the attention mechanism is as follows: 

\begin{equation}\label{eq4}
	\begin{aligned}
		& B_C = \text{Cat}(B_{F,T}, B_{F,T-1}, \dots, B_{F,T-N}), \\
		& \mathrm{SA} = \mathrm{Conv}{1\times1}(\mathrm{Conv}_{D-DW}(\mathrm{Conv}_{DW}(B_C))), \\
	\end{aligned}
\end{equation}

where $B_C$ represents the multi-frame BEV features, and $\mathrm{SA}$ denotes the static attention.
By using these methods, the attention mechanism is able to effectively model a large receptive field, identifying subtle relationships between elements in the frame. This not only aids in extracting local features but also enables the capture of global information across the image or point cloud. This mechanism reduces inherent noise in the data, improving the model's robustness, especially in static scenes with complex spatial structures. The ability to identify long-range dependencies and subtle patterns enhances detection performance, making it particularly effective in handling challenging static environments.

\begin{figure}[h]
	
	\centering
	\includegraphics[width=1.0\linewidth]{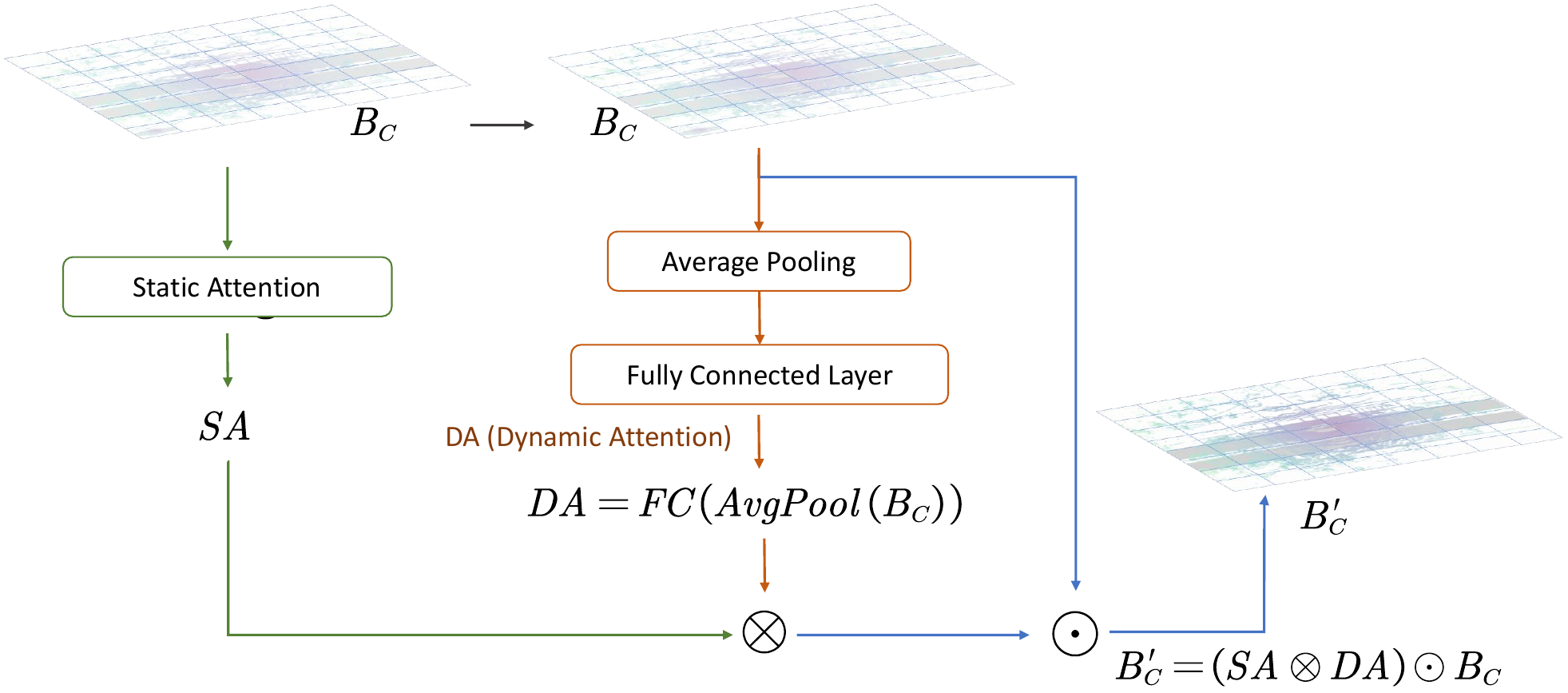}	
	\caption{Intra-frame Static Attention.						
	}
	\label{inter-f}
\end{figure}

\begin{figure*}[t]

	\centering
	\includegraphics[width=0.78\linewidth]{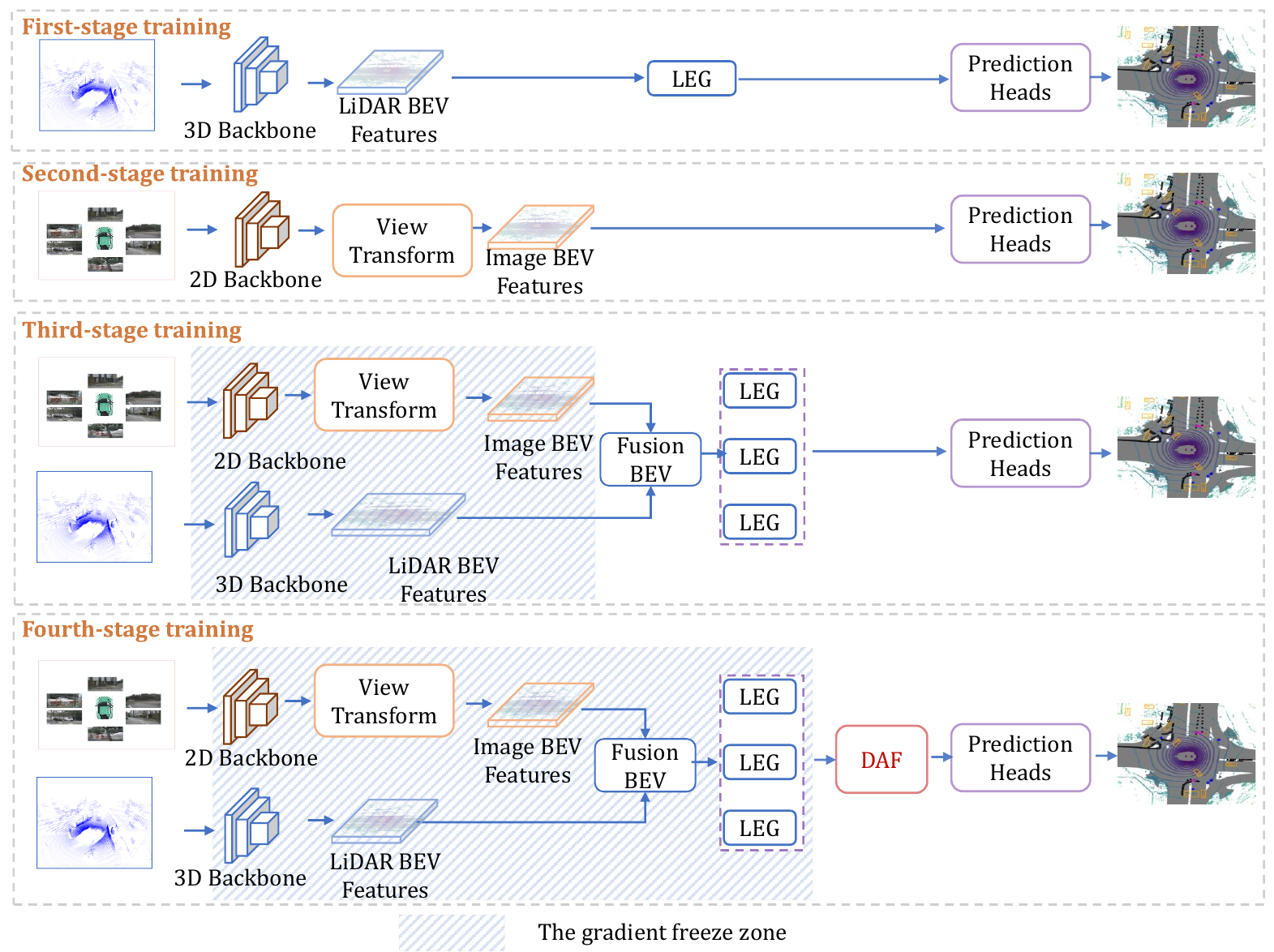}
	\caption{\centering Four-Stage Training Method.}
	\label{fourStage}
\end{figure*}
\textbf{Inter-frame Dynamic Attention:} Unlike static attention, inter-frame dynamic attention is designed to capture temporal variations and evolution across frames. This mechanism uses a squeeze-and-excitation approach to learn dynamic attention weights for each channel, enabling the model to adjust the importance of channels in response to temporal changes. The specific formula is as follows:

\begin{equation}\label{eq4}
	\mathrm{DA} = \mathrm{FC}(\mathrm{AvgPool}(B_C)),
\end{equation}

where $\mathrm{DA}$ represents the dynamic attention, 
$\mathrm{FC}(\cdot)$ denotes a fully connected layer, and $\mathrm{AvgPool}(\cdot)$ refers to the average pooling layer.

By combining dynamic attention with static attention, the model is enabled to capture both temporal evolution and spatial dependencies. The specific formula is as follows:

\begin{equation}\label{eq4}
	B_C^{\prime} = (\mathrm{SA} \otimes \mathrm{DA}) \odot B_C,
\end{equation}

where $\otimes$ denotes the Kronecker product between static attention (SA)  and dynamic attention (DA), and $\odot$ represents the Hadamard product between the attention-weighted features  and the original BEV feature map $B_C$. The resulting $B_C^{\prime}$ represents the optimized BEV features.
This dynamic weighting makes the model more sensitive to changes in the scene, particularly when detecting and tracking moving objects in sequential data, thus enhancing its ability to handle temporal variations in the input sequence.

\subsection{Training Method }
In dual-branch multi-modal fusion, naively optimizing all parameters jointly often leads to instability and suboptimal performance. 
This is because multi-modal temporal modeling spans three coupled optimization subspaces: intra-modal feature learning, cross-modal fusion, and cross-frame temporal relations. 
If all parameters are updated simultaneously, gradients from different modalities or temporal frames may conflict, resulting in oscillations or degraded convergence. To mitigate these issues, Co-Fusion4D introduces a progressive four-stage training framework (Fig.~\ref{fourStage}), inspired by the cooperative optimization philosophy of Co-Fix3D. 
The core idea is to \textit{decouple} the optimization subspaces across stages, allowing each to converge under controlled conditions before subsequent coupling. 
This design ensures stable feature representations prior to introducing temporal modeling and avoids the gradient interference inherent in naive joint optimization.

\subsubsection*{Stage 1--3: Progressive single-frame and multi-modal optimization}
The first three stages follow the Co-Fix3D approach, with the following objectives:

\begin{enumerate}
	\item Stage 1: LiDAR branch pretraining. The point cloud encoder is trained independently, ensuring it learns stable geometric representations without interference from the image modality.
	\item Stage 2: Image branch pretraining. With the LiDAR branch frozen, the image branch is trained to map 2D semantic features to 3D BEV space, stabilizing the image modality representation before fusion.
	\item Stage 3: Multi-modal fusion training. Once each modality has stabilized, cross-modal fusion modules are trained. This ensures that multi-modal feature integration occurs on robust, pre-converged representations, effectively mitigating modality bias propagation.
\end{enumerate}

By decoupling modality-specific learning in the first three stages, Co-Fusion4D guarantees that each single-frame representation reaches sufficient stability before temporal modeling is introduced.

\subsubsection*{Stage 4: Temporal multi-frame training with protective freezing}
The fourth stage introduces multi-frame temporal learning. 
Critically, both the LiDAR and image branches are \textbf{frozen}, and only the alignment modules, DAF, and detection heads are updated. 
This \textit{protective freezing} strategy addresses two key challenges:

\begin{itemize}
	\item \textbf{Gradient interference mitigation:} Without freezing, temporal loss gradients would propagate into single-frame encoders, destabilizing previously learned representations due to residual misalignment or semantic drift in historical frames.
	\item \textbf{Focused temporal optimization:} Freezing single-frame features restricts the optimization space to cross-frame relationships, allowing alignment modules to compensate for spatial shifts, DAF to learn selective temporal weighting, and detection heads to operate on fused spatiotemporal features without disrupting stable single-frame embeddings.
\end{itemize}

This design ensures that temporal learning benefits from robust, pre-converged features, maximizing DAF's selective fusion capabilities within a constrained optimization space.

\subsubsection*{Comparison with existing methods}
Co-Fusion4D differs fundamentally from prior approaches in two aspects:

\begin{itemize}
	\item \textbf{Modality-decoupled single-frame training:} Unlike BEVFusion4D and GAFusion, which jointly optimize LiDAR and image branches after LiDAR pretraining, Co-Fusion4D separates LiDAR and image pretraining to avoid modality gradient conflicts and ensure stable feature representations before fusion.
	\item \textbf{Protected temporal optimization:} Previous multi-frame methods either allow encoder parameters to update (BEVFusion4D) or aggregate historical features without explicit protection (GAFusion), potentially destabilizing single-frame representations. In contrast, Co-Fusion4D freezes all single-frame parameters, isolating the temporal optimization space.
\end{itemize}

In summary, the four-stage training strategy is not a mere extension of stage count but embodies two design principles: 
(1) decoupled single-frame modality training to suppress modality bias propagation; 
(2) protective freezing in temporal training to confine optimization to cross-frame relationships. 
The synergy between this training strategy and DAF is critical: frozen single-frame features provide a stable foundation, enabling DAF to learn meaningful spatiotemporal attention distributions, which maximally leverages the constrained temporal optimization space.

\section{Experiments}
\subsection{Datasets }	
We conduct  comprehensive evaluations using the widely recognized nuScenes dataset \cite{caesar2020nuscenes}, a standard benchmark for various autonomous driving tasks, including 3D detection, segmentation, and tracking. This large-scale dataset comprises 1,000 varied scenarios, featuring data from six cameras and point clouds captured by a 32-beam LiDAR sensor. The camera setup provides a 360-degree field of view, perfectly synchronized with the LiDAR coverage, making it well-suited for evaluating LiDAR-camera fusion algorithms. Additionally, the nuScenes dataset is split into training, validation, and test subsets, with 700, 150, and 150 scenarios, respectively. Performance assessments are carried out using official metrics, including Mean Average Precision (mAP) and the nuScenes Detection Score (NDS).

\subsection{Implementation Details}

Our model is developed on the PyTorch platform \cite{paszke2017automatic} and leverages the community-driven MMDetection3D framework \cite{contributors2020mmdetection3d}. The camera branch utilizes a Swin Transformer as the backbone, consisting of heads with 3, 6, 12, and 24 layers. The input image is resized and cropped to $386 \times 1056$ pixels to standardize the data for processing. In the Lift-Splat-Shoot (LSS) configuration, the frustum ranges are set as follows: X coordinates range from [-54m, 54m] with a step of 0.6m, Y coordinates follow the same range, and Z coordinates are set from [-5m, 3m] with a 0.6m step. The depth range spans from 1m to 60m with a step of 0.5m. The BEV grid resolution is configured at $180 \times 180$, which is aligned with the $8 \times$ downsampled BEV features generated by VoxelNet \cite{zhou2018voxelnet}, with a channel count of 128. The model was trained on 4 Nvidia 4090 GPUs with a total batch size of 8. We used the AdamW optimizer with an initial learning rate of $1.0 \times 10^{-4}$. A one-cycle learning rate policy was applied to dynamically adjust the learning rate during training, enhancing convergence speed and overall efficiency.
\subsection{ 3D Object Detection Results}
\begin{table*}[htpb]
	\centering
	\caption{\textbf{3D Object Detection Performance on the nuScenes test set.} The notion of modality: Camera (C), LiDAR (L) and Temporal (T). `C.V.', `T.L.', `B.R.', `M.T.', `Ped.', and ‘T.C.’ indicate the construction vehicle, trailer, barrier, motorcycle, pedestrian, and traffic cone, respectively. The best results in each column are marked in bold font.}
	\resizebox{\linewidth}{!}{
	\begin{tabular}{l|c|cc|ccccccccccc}
		\toprule
		\textbf{Method}   &   \textbf{Modality}  &  \textbf{mAP}  &  \textbf{NDS}  & \textbf{Car} & \textbf{Truck} &  \textbf{C.V.} &  \textbf{Bus} &  \textbf{T.L.} &  \textbf{B.R.} & \textbf {M.T.} &  \textbf{Bike} &  \textbf{Ped.} &  \textbf{T.C.} \\ \midrule \midrule
		TransFusion-L~\cite{bai2022transfusion}  &L         & 65.5       & 70.2               & 86.2         & 56.7           & 28.2          & 66.3         & 58.8             & 78.2             & 68.3           & 44.2          & 86.1          & 82.0          \\ 
 		LiDARMultiNet~\cite{ye2023lidarmultinet}   & L                 & 71.6         & 67.0         & 86.9         & 57.4           & 31.5         &64.7         & 61.0             & 73.5             & 75.3           & 47.6         & 87.2          & 85.1         \\ 
		FocalFormer3D~\cite{chen2023focalformer3d} &L    & 68.7               & 72.6             & 87.2         & 57.1           & 34.4          & 69.6         & 64.9             & 77.8             & 76.2           & 49.6        & 88.2         & 82.3         \\ 
		
		\midrule
		
		TransFusion-LC~\cite{bai2022transfusion} &LC      & 68.9        & 71.7               & 87.1         & 60.0           & 33.1          & 68.3         & 60.8             & 78.1             & 73.6           & 52.9          & 88.4          & 86.7          \\
		BEVFusion~\cite{liang2022bevfusion} &LC       & 69.2         & 71.8             & 88.1         & 60.9           & 34.4          & 69.3         & 62.1             & 78.2             & 72.2           & 52.2          & 89.2          & 85.5          \\
		BEVFusion~\cite{liu2023bevfusion}&LC       & 70.2           & 72.9             & 88.6         & 60.1           & 39.3          & 69.8         & 63.8             & 80.0             & 74.1           & 51.0          & 89.2          & 86.5          \\  

		ObjectFusion~\cite{cai2023objectfusion} &LC    & 71.0  & 73.3              &    89.4          &      59.0          &      40.5         &      71.8        &    63.1              &       80.0           &      78.1          &  53.2             &      90.7         &   87.7        \\
		MSMDFusion~\cite{jiao2023msmdfusion} &LC    & 71.5  & 74.0              &    88.4          &      61.0          &      35.2         &      71.4        &    64.2              &       80.7           &      76.9          &  58.3             &      90.6         &   88.1        \\
		SparseFusion~\cite{xie2023sparsefusion} &LC  & 72.0              & 73.8                 & 88.0         & 60.2           & 38.7          & 72.0         & 64.9             & 79.2             & 78.5           & 59.8          & 90.9          & 87.9          \\
		FocalFormer3D~\cite{chen2023focalformer3d}  &LC     & 71.6           & 73.9               & 88.5         & 61.4           & 35.9          & 71.7         & 66.4             & 79.3             & 80.3           & 57.1          & 89.7          & 85.3          \\
		IS-Fusion~\cite{yin2024fusion} &LC     & {73.0}          & 75.2              & 88.3             & 62.7               & {38.4}              &   74.9         &     {67.3}             &       78.1           &   82.4         &    59.5             &   89.3           &   \textbf{ 89.2}           \\
		
		GraphBEV~\cite{song2025graphbev} &LC     & {71.7}          & 73.6              &  89.2             & 60.0              & 40.8           &  72.1          &     64.5           &      80.1
		&   76.8       &    53.3          &   90.9         &   88.9         \\
		Co-Fix3D~\cite{li2024cofix3d}
		&LC   &  72.3   & 74.7              & 89.6           &60.8        & 37.9              &   73.3           &     65.2            &       80.0        &   77.6         &    62.7           &  88.7          &   86.7         \\
		MTA~\cite{lin2025method}  &LC     & {71.8}          & 73.7              &  88.6             & 63.3             & 38.4           &  71.9       &     64.7           &      79.4
		&  75.5       &    57.2       &   90.8        &   88.5         \\	
	    \midrule
		LIFT~\cite{zeng2022lift}&LC     & 65.1        & 70.2         & 87.7      &   55.1      & 29.4           & 62.4       &   59.3     &   69.3     &    70.8   &    47.7         &   86.1    &  83.2           \\
		
		BEVFusion4D~\cite{cai2023bevfusion4d}&LC    & 73.3         & 74.7           &  89.7       & 65.6         & 41.1          & 72.9        &   66.0        &   \textbf{81.0}         &    79.5     &    58.6          &   90.9    &  87.7           \\
		GAFusion~\cite{li2024gafusion}  &LC     & {73.6}          & 74.9              &  89.4             & 65.3             & \textbf{42.4}            &  73.7          &     65.8           &      79.2          &    80.8        &    60.2            &   \textbf{92.3}           &    87.0            \\

		Co-Fusion4D (Our)
		&LC    &  \textbf{74.9}   & \textbf{75.6}               & \textbf{90.5}            &\textbf{65.7}           & 39.7             &   \textbf{77.1 }          &     \textbf{69.9}             &      \textbf{81.0}         &   \textbf{ 83.5}          &    \textbf{63.1}            &   91.2         &    88.2           \\
		\bottomrule	
	\end{tabular}
}
	\label{tb:test}
\end{table*}

\textbf{3D Object Detection on the Test Dataset:} We evaluate the performance of Co-Fusion4D against leading LiDAR-based (denoted as 'L') and  multimodal (denoted as 'LC') 3D object detectors on the nuScenes test set. Co-Fusion4D demonstrates superior performance compared to existing state-of-the-art (SOTA) 3D object detection algorithms. As shown in Table~\ref{tb:test}, Co-Fusion4D outperforms the baseline Co-Fix3D, improving the mean Average Precision (mAP) by 2.0\% and the nuScenes Detection Score (NDS) by 1.1\%. When compared to other recent multimodal detection methods, such as IS-Fusion, GraphBEV, and GAFusion, Co-Fusion4D continues to lead, with mAP improvements of 1.3\%, 2.6\%, and 0.7\%, respectively, and corresponding NDS gains of 0.6\%, 2.2\%, and 0.9\%. Additionally, Co-Fusion4D achieves the highest detection performance in specific categories, such as Car and Bus, emphasizing that multi-frame fusion significantly enhances the model's ability to detect moving objects.

	\begin{table}[h]
	\centering
	\caption{Performance comparison on the nuScenes validation set.}
	\resizebox{0.48\textwidth}{!}
	{
		\begin{tabular}{l|l|ll}
			\toprule
			Method  & \begin{tabular}[c]{@{}l@{}}Image\\ Encoder\end{tabular} & mAP  & NDS\\ \midrule \midrule
			TransFusion-LC~\cite{bai2022transfusion}     & 	ResNet-50                              & 67.5                        & 71.3    \\
			BEVFusion~\cite{liang2022bevfusion}                     & Swin-T                     & 68.5                               &   71.4       \\

			SparseFusion~\cite{xie2023sparsefusion}       & ResNet-50                                & 71.0                         & 73.1   \\
			IS-Fusion~\cite{yin2024fusion}                & Swin-T                              & 72.8                        & 74.0   
			\\
			Co-Fix3D~\cite{li2024cofix3d}            & ResNet-50                             & 70.8                      & 73.6 
			\\
			BEVFusion4D~\cite{cai2023bevfusion4d}                & Swin-T                              & 72.0                        & 73.5  
			\\
			GAFusion~\cite{li2024gafusion}                 & Swin-T                         & 72.1                      & 73.5   
			\\

			\midrule
			Co-Fusion4D (Ours) & Swin-T                           & \textbf{73.9}                          &  \textbf{75.3} \\ 
			\bottomrule
		\end{tabular}
	}
	
	\label{tb:val}
\end{table}

\textbf{3D object detection on val set.} We also conduct experiments on the validation set, as shown in Table~\ref{tb:val}. Co-Fusion4D demonstrates superior performance compared to existing state-of-the-art (SOTA) 3D object detection algorithms. The results reveal that Co-Fusion4D outperforms the baseline Co-Fix3D, improving the  mAP by 2.1\% and the NDS) by 1.7\%. When compared to other recent multimodal detection methods such as IS-Fusion, BEVFusion4D, and GAFusion, Co-Fusion4D still leads, with mAP improvements of 1.1\%, 1.9\%, and 1.8\%, respectively, and corresponding NDS gains of 1.3\%, 1.8\%, and 1.8\%. These results strongly demonstrate that our DAF module effectively integrates multi-frame information and, combined with our proposed training approach, harnesses the complementary strengths of point cloud and image modalities, avoiding underfitting and ensuring optimal performance.

\begin{figure*}[h!]
	\centering
	\includegraphics[width=0.95\linewidth]{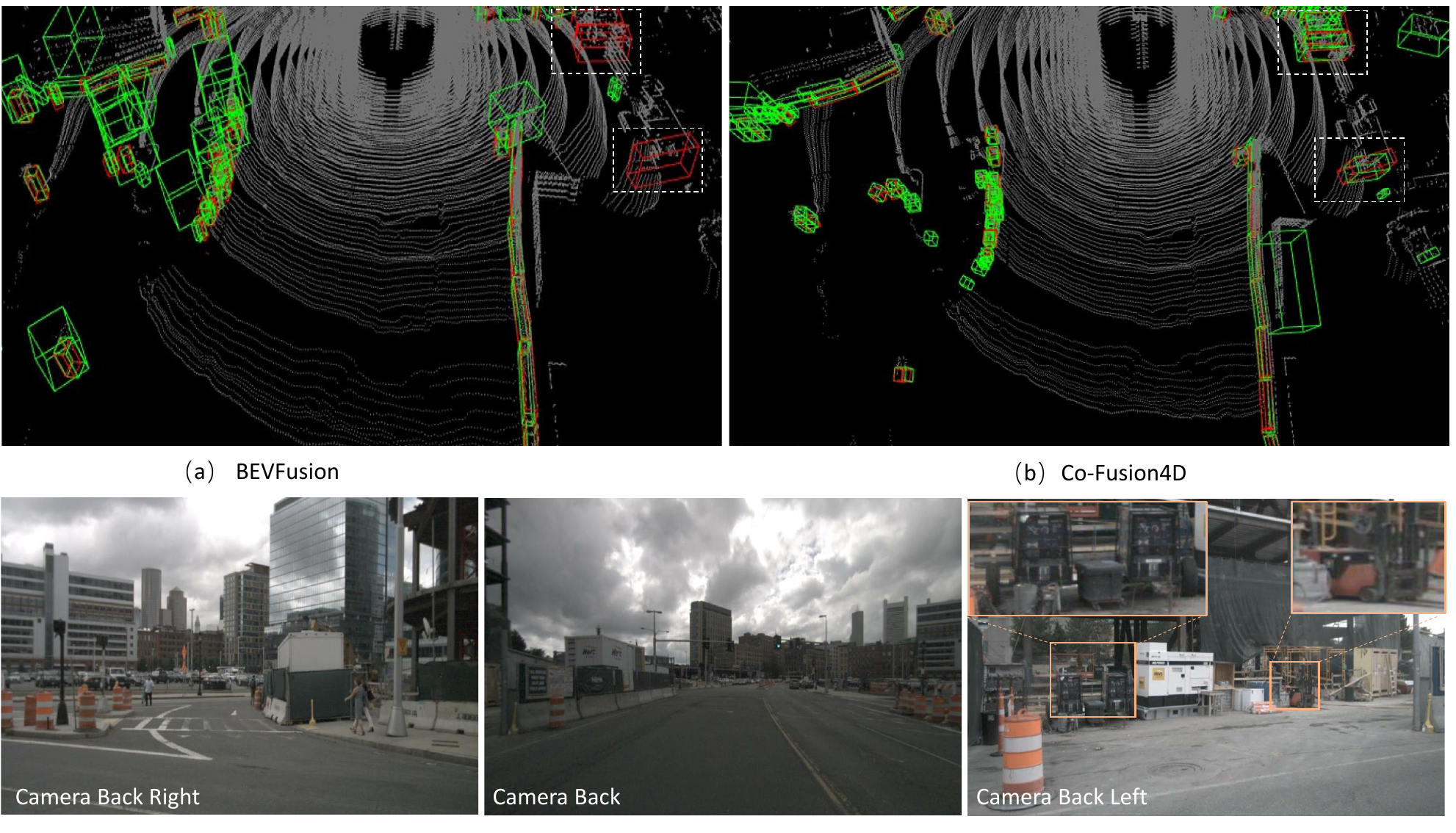}	
	\caption{3D object detection results on the nuScenes validation dataset. In the point cloud image on the far right, red boxes represent the ground truth (GT), while blue boxes indicate the predicted detections. A total of 100 bounding boxes are shown.					
	}
	\label{viz}
	\vspace*{-4mm}
\end{figure*}

\subsection{Ablation Studies}
\label{sec:ablation_cofusion4d}

We conduct ablation studies on the nuScenes validation set to isolate the contribution of each key design in Co-Fusion4D. 
Due to strong interdependence among temporal components, we follow a progressive enabling protocol and report mAP/NDS.

\begin{table}[h]
	\centering
	\caption{Multi-frame pathway and alignment necessity on nuScenes validation.}
	\label{tab:path_alignment_ablation}
	\begin{tabular}{cccccc}
		\toprule
		Hist. PCD & PCD Align & Hist. BEV & BEV Align & mAP & NDS \\
		\midrule
		&  &  &  & 73.1 & 74.2 \\
		$\checkmark$ &  &  &  & 73.1 & 74.3 \\
		$\checkmark$ & $\checkmark$ &  &  & 73.4 & 74.7 \\
		$\checkmark$ & $\checkmark$ & $\checkmark$ &  & 73.5 & 74.9 \\
		$\checkmark$ & $\checkmark$ & $\checkmark$ & $\checkmark$ & \textbf{73.9} & \textbf{75.3} \\
		\bottomrule
	\end{tabular}
\end{table}

\begin{table*}[h]
	\centering
	\caption{Ablation on training strategies.}
	\label{tab:training_strategy_ablation}
	
	\begin{tabular}{cp{10.2cm}cc}
		\toprule
		ID & Description & mAP & NDS \\
		\midrule
		(a) & Multi-frame from Stage-1 (symmetric fusion) & 61.3 & 67.8 \\
		(b) & Multi-frame in Stage-2 (symmetric fusion) & 63.9 & 68.9 \\
		(c) & Multi-frame in Stage-3 (symmetric fusion) & 69.2 & 71.8 \\
		(d) & Skip Stage-2; multi-frame after Stage-3 (symmetric fusion) & 66.4 & 70.3 \\
		(e) & Stage-4: freeze encoders + Cat (w/o DAF) & 73.3 & 74.6 \\
		(f) & Stage-4: symmetric fusion + DAF (w/o freezing) & 71.1 & 73.3 \\
		(g) & Stage-4: asymmetric fusion + DAF (with freezing) & \textbf{73.9} & \textbf{75.3} \\
		\bottomrule
	\end{tabular}
\end{table*}

\begin{table*}[h]
	\centering
	\caption{Category-wise analysis on dynamic classes with and without historical frames on the nuScenes validation set.}
	\label{tab:dynamic_class_ablation}
	\begin{tabular}{c c c c c c c c}
		\hline
		History & mAP & NDS & Car & Truck & C.V. & M.T. & Bike \\
		\hline
		& 73.1 & 74.2 & 91.1 & 69.2 & 35.4 & 85.1 & 74.6 \\
		$\checkmark$ & \textbf{73.9} & \textbf{75.3} & 91.6 & 70.5 & 36.8 & 86.6 & 76.6 \\
		\hline
		Gain & +0.8 & +1.1 & +0.5 & +1.3 & +1.4 & +1.5 & +2.0 \\
		\hline
	\end{tabular}
\end{table*}

\paragraph{Alignment is necessary for effective temporal aggregation.}
Table~\ref{tab:path_alignment_ablation} evaluates the multi-frame pathway by incrementally enabling historical cues and alignment.
Simply adding historical point clouds without alignment yields negligible gains, indicating that temporal evidence is not directly exploitable under cross-frame inconsistencies.
Geometric alignment improves performance, suggesting it is a prerequisite for multi-frame utility.
Adding historical BEV features provides additional gains but remains limited without semantic-level alignment.
Enabling BEV feature alignment achieves the best results, supporting the need for a constrained route from geometric to semantic alignment.

\paragraph{DAF outperforms generic fusion operators under a frozen-encoder protocol.}
Table~\ref{tab:fusion_strategy_ablation} compares spatio-temporal fusion strategies under Stage-4 training where the single-frame encoder is frozen and only fusion modules are optimized.
Direct fusion (Add) underperforms, while concatenation (Cat) is stronger, implying that early mixing can be harmful under temporal misalignment.
Channel attention (SE/ECA) brings limited improvements, suggesting channel-only recalibration is insufficient for spatially variant temporal quality.
Cross-attention improves NDS but is costly due to quadratic token interactions w.r.t. BEV resolution.
DAF achieves the best performance by decoupling intra-frame spatial aggregation and inter-frame temporal reweighting.

\begin{table}[h]
	\centering

	\caption{Fusion strategy comparison on nuScenes validation (Stage-4, frozen single-frame encoder).}
	\label{tab:fusion_strategy_ablation}
		\resizebox{0.48\textwidth}{!}
{	\begin{tabular}{llcc}
		\toprule
		Category & Method & mAP & NDS \\
		\midrule
		Direct fusion & Add & 73.2 & 74.2 \\
		Direct fusion & Cat & 73.4 & 74.6 \\
		Channel attention & Cat+SE & 73.6 & 74.8 \\
		Channel attention & Cat+ECA & 73.5 & 74.7 \\
		Cross-frame attention & Cross-Attn (Cur=Q, Hist=K/V) & 73.6 & 74.9 \\
		Spatio-temporal co-attention & DAF & \textbf{73.9} & \textbf{75.3} \\
		\bottomrule
	\end{tabular}}

\end{table}

\begin{table}[h]
	\centering
	\caption{Component ablation of DAF on nuScenes validation (Cat baseline in the first row).}
	\label{tab:daf_component_ablation}
	\begin{tabular}{cccccc}
		\toprule
		Intra-frame & Inter-frame & mAP & NDS & $\Delta$mAP & $\Delta$NDS \\
		\midrule
		&  & 73.4 & 74.6 & -- & -- \\
		$\checkmark$ &  & 73.3 & 74.3 & $-0.1$ & $-0.3$ \\
		& $\checkmark$ & 73.2 & 74.1 & $-0.2$ & $-0.5$ \\
		$\checkmark$ & $\checkmark$ & \textbf{73.9} & \textbf{75.3} & $+0.5$ & $+0.7$ \\
		\bottomrule
	\end{tabular}
\end{table}

\paragraph{DAF requires joint static--dynamic attention to be effective.}
Table~\ref{tab:daf_component_ablation} ablates DAF into intra-frame static attention and inter-frame dynamic attention.
Either component alone is inferior to Cat, while the combination yields the best result.
This indicates a non-additive coupling: static attention without temporal gating may amplify historical noise, whereas dynamic attention without strong spatial structure may produce unreliable temporal weights.
The joint design enables discriminative spatial representations and stable temporal reweighting.

\begin{table}[h]
	\centering
	\caption{Impact of the number of frames on detection performance.}
	\label{tab:number_of_frames_ablation}
	\begin{tabular}{ccccc}
		\toprule
		\#Frames & mAP & NDS & $\Delta$mAP & $\Delta$NDS \\
		\midrule
		1 & 73.1 & 74.2 & -- & -- \\
		3 & 73.9 & 75.3 & $+0.8$ & $+1.1$ \\
		5 & 74.0 & 75.4 & $+0.9$ & $+1.2$ \\
		7 & 74.0 & 75.3 & $+0.9$ & $+1.1$ \\
		\bottomrule
	\end{tabular}
\end{table}

\paragraph{Temporal window length: gains saturate beyond short history.}
Table~\ref{tab:number_of_frames_ablation} shows that expanding to 3 frames yields the largest gain, while improvements saturate beyond 3 frames.
Longer history can introduce accumulated alignment residuals and scene changes, leading to marginal or even slightly negative returns.
We therefore adopt 3 frames as a default trade-off between accuracy and efficiency.

\paragraph{Category-wise gains concentrate on highly dynamic classes.}
Table~\ref{tab:dynamic_class_ablation} reports per-class performance with/without history.
Temporal modeling yields modest gains for well-observed and stable classes (e.g., car/truck) and larger gains for more dynamic or unstable classes (e.g., construction vehicle/motorcycle/bicycle).
No degradation is observed among the listed dynamic categories, indicating that current-frame-centric temporal modeling suppresses harmful temporal interference.

\paragraph{Distance robustness: temporal fusion primarily benefits far-range objects.}
Table~\ref{tab:distance_robustness_ablation} shows that improvements are concentrated in the 30--50m range, where single-frame sensing is limited by LiDAR sparsity and camera depth uncertainty.
Near- and mid-range gains are marginal, consistent with sufficient single-frame evidence at shorter distances.
This supports temporal accumulation as a targeted remedy for far-range perception.

\begin{table}[h]
	\centering
	\caption{Robustness w.r.t. distance (mAP, \%).}
	\label{tab:distance_robustness_ablation}
	\begin{tabular}{cccc}
		\toprule
		\#Frames & 0--15m & 15--30m & 30--50m \\
		\midrule
		1 & 76.5 & 69.3 & 55.6 \\
		3 & 76.9 & 70.6 & 58.9 \\
		5 & 76.9 & 70.7 & 59.3 \\
		\bottomrule
	\end{tabular}
\end{table}

\begin{table}[h]
	\centering
	\caption{Latency breakdown (ms) and scalability w.r.t. number of frames.}
	\label{tab:latency_breakdown_ablation}
	\resizebox{0.48\textwidth}{!}
	{
		\begin{tabular}{lcccc}
			\toprule
			Module & 1 frame & 2 frames & 3 frames & 4 frames \\
			\midrule
			LiDAR backbone & 68 & 96 & 134 & 172 \\
			Image backbone (Swin-T) & 74 & 132 & 193 & 241 \\
			View transform (LSS) & 53 & 121 & 192 & 263 \\
			PCD align + BEV align & -- & 1 & 1 & 1 \\
			DAF & 2 & 2 & 3 & 4 \\
			Others & 69 & 81 & 103 & 115 \\
			\midrule
			Total & 266 & 433 & 626 & 796 \\
			\bottomrule
		\end{tabular}
	}
\end{table}

\paragraph{Training strategy: late temporal learning with encoder freezing is critical.}
Table~\ref{tab:training_strategy_ablation} ablates when temporal learning is introduced and whether the encoder is frozen.
Early multi-frame training with symmetric fusion is suboptimal, likely due to unstable single-frame representations being corrupted by temporal noise.
Stage-4 freezing yields a strong baseline even with Cat fusion, indicating that protecting converged representations stabilizes temporal optimization.
Training DAF without freezing underperforms, suggesting temporal reweighting becomes unstable when features drift.
The best configuration is Stage-4 with encoder freezing and asymmetric fusion with DAF.

\paragraph{Latency breakdown: temporal fusion is lightweight.}
Table~\ref{tab:latency_breakdown_ablation} decomposes inference latency on a single NVIDIA 4090D GPU.
Latency increases approximately linearly with the number of frames, while alignment and DAF contribute only a few milliseconds.
The major overhead comes from the image pipeline, especially multi-frame view transformation (LSS).
This suggests future acceleration should prioritize reducing multi-frame view transformation cost.

As shown in Fig.~\ref{viz}, Co-Fusion4D efficiently detects moving objects, particularly cars and buses, after applying multi-frame fusion. This underscores the superior detection performance of our model.

\section{Conclusions}
This paper proposed a temporal BEV-based multimodal 3D object detection method, featuring a novel four-stage training strategy that effectively addresses the common issue of overfitting in BEV-based multimodal models. It employs a four-stage training method, with Co-Fusion4D effectively integrating these modalities to maximize detection performance. Additionally, Co-Fusion4D introduces the Dual Attention Fusion (DAF) module, which efficiently integrates multi-frame BEV features, significantly improving the model's ability to detect and track dynamic objects. By combining novel training strategies and feature fusion methods,  Co-Fusion4D demonstrates superior performance in 3D object detection, particularly in dynamic and complex environments. This work not only advances multimodal 3D detection but also provides valuable insights for future research aimed at improving object detection in dynamic settings.

%
%
%
{			
	\bibliographystyle{unsrt}
	\bibliography{root}

@String(AAAI = {AAAI})

@article{yan2018second,
	title={Second: Sparsely embedded convolutional detection},
	author={Yan, Yan and Mao, Yuxing and Li, Bo},
	journal={Sensors},
	volume={18},
	number={10},
	pages={3337},
	year={2018},
	publisher={MDPI}
}

@inproceedings{shi2020pv,
	title={Pv-rcnn: Point-voxel feature set abstraction for 3d object detection},
	author={Shi, Shaoshuai and Guo, Chaoxu and Jiang, Li and Wang, Zhe and Shi, Jianping and Wang, Xiaogang and Li, Hongsheng},
	booktitle={Proceedings of the IEEE/CVF conference on computer vision and pattern recognition},
	pages={10529--10538},
	year={2020}
}

@inproceedings{qi2017pointnet,
	title={Pointnet: Deep learning on point sets for 3d classification and segmentation},
	author={Qi, Charles R and Su, Hao and Mo, Kaichun and Guibas, Leonidas J},
	booktitle={Proceedings of the IEEE conference on computer vision and pattern recognition},
	pages={652--660},
	year={2017}
}

@inproceedings{shi2019pointrcnn,
	title={Pointrcnn: 3d object proposal generation and detection from point cloud},
	author={Shi, Shaoshuai and Wang, Xiaogang and Li, Hongsheng},
	booktitle={Proceedings of the IEEE/CVF conference on computer vision and pattern recognition},
	pages={770--779},
	year={2019}
}

@inproceedings{caesar2020nuscenes,
	title={nuscenes: A multimodal dataset for autonomous driving},
	author={Caesar, Holger and Bankiti, Varun and Lang, Alex H and Vora, Sourabh and Liong, Venice Erin and Xu, Qiang and Krishnan, Anush and Pan, Yu and Baldan, Giancarlo and Beijbom, Oscar},
	booktitle={Proceedings of the IEEE/CVF conference on computer vision and pattern recognition},
	pages={11621--11631},
	year={2020}
}

@inproceedings{li2022bevformer,
	title={Bevformer: Learning bird’s-eye-view representation from multi-camera images via spatiotemporal transformers},
	author={Li, Zhiqi and Wang, Wenhai and Li, Hongyang and Xie, Enze and Sima, Chonghao and Lu, Tong and Qiao, Yu and Dai, Jifeng},
	booktitle={European conference on computer vision},
	pages={1--18},
	year={2022},
	organization={Springer}
}

@article{wan2025focalfusion,
  title={FocalFusion: An object-centric temporal fusion framework for multi-modal 3D detection},
  author={Wan, Yuting and Sun, Liguo and Hao, Jiuwu and Lv, Pin},
  journal={Neurocomputing},
  pages={131914},
  year={2025},
  publisher={Elsevier}
}

@inproceedings{wang2023exploring,
  title={Exploring object-centric temporal modeling for efficient multi-view 3d object detection},
  author={Wang, Shihao and Liu, Yingfei and Wang, Tiancai and Li, Ying and Zhang, Xiangyu},
  booktitle={Proceedings of the IEEE/CVF international conference on computer vision},
  pages={3621--3631},
  year={2023}
}

@inproceedings{singh2023transformer,
  title={Transformer-based sensor fusion for autonomous driving: A survey},
  author={Singh, Apoorv},
  booktitle={Proceedings of the IEEE/CVF international conference on computer vision},
  pages={3312--3317},
  year={2023}
}

@article{li2025bevfix,
  title={BEVFix: Deep feature enhancement for robust 3D object detection},
  author={Li, Wenxuan and Zhou, Jian and Chen, Chi and Yu, Hongkai and Du, Bo and Zou, Qin},
  journal={Neural Networks},
  volume={190},
  pages={107675},
  year={2025},
  publisher={Elsevier}
}

@article{ma2024vision,
  title={Vision-centric bev perception: A survey},
  author={Ma, Yuexin and Wang, Tai and Bai, Xuyang and Yang, Huitong and Hou, Yuenan and Wang, Yaming and Qiao, Yu and Yang, Ruigang and Zhu, Xinge},
  journal={IEEE Transactions on Pattern Analysis and Machine Intelligence},
  volume={46},
  number={12},
  pages={10978--10997},
  year={2024},
  publisher={IEEE}
}

@article{qi2017pointnet++,
  title={Pointnet++: Deep hierarchical feature learning on point sets in a metric space},
  author={Qi, Charles Ruizhongtai and Yi, Li and Su, Hao and Guibas, Leonidas J},
  journal={Advances in neural information processing systems},
  volume={30},
  year={2017}
}

@article{paszke2017automatic,
	title={Automatic differentiation in pytorch},
	author={Paszke, Adam and Gross, Sam and Chintala, Soumith and Chanan, Gregory and Yang, Edward and DeVito, Zachary and Lin, Zeming and Desmaison, Alban and Antiga, Luca and Lerer, Adam},
	year={2017}
}

@misc{contributors2020mmdetection3d,
	title={MMDetection3D: OpenMMLab next-generation platform for general 3D object detection},
	author={Contributors, MMDetection3D},
	year={2020}
}

@inproceedings{zhou2018voxelnet,
	title={Voxelnet: End-to-end learning for point cloud based 3d object detection},
	author={Zhou, Yin and Tuzel, Oncel},
	booktitle={Proceedings of the IEEE conference on computer vision and pattern recognition},
	pages={4490--4499},
	year={2018}
}

@inproceedings{bai2022transfusion,
  title={Transfusion: Robust lidar-camera fusion for 3d object detection with transformers},
  author={Bai, Xuyang and Hu, Zeyu and Zhu, Xinge and Huang, Qingqiu and Chen, Yilun and Fu, Hongbo and Tai, Chiew-Lan},
  booktitle={Proceedings of the IEEE/CVF conference on computer vision and pattern recognition},
  pages={1090--1099},
  year={2022}
}

@inproceedings{ye2023lidarmultinet,
  title={Lidarmultinet: Towards a unified multi-task network for lidar perception},
  author={Ye, Dongqiangzi and Zhou, Zixiang and Chen, Weijia and Xie, Yufei and Wang, Yu and Wang, Panqu and Foroosh, Hassan},
  booktitle={Proceedings of the AAAI Conference on Artificial Intelligence},
  volume={37},
  number={3},
  pages={3231--3240},
  year={2023}
}

@inproceedings{chen2022deformable,
  title={Deformable feature aggregation for dynamic multi-modal 3D object detection},
  author={Chen, Zehui and Li, Zhenyu and Zhang, Shiquan and Fang, Liangji and Jiang, Qinhong and Zhao, Feng},
  booktitle={European conference on computer vision},
  pages={628--644},
  year={2022},
  organization={Springer}
}

@article{li2022unifying,
  title={Unifying voxel-based representation with transformer for 3d object detection},
  author={Li, Yanwei and Chen, Yilun and Qi, Xiaojuan and Li, Zeming and Sun, Jian and Jia, Jiaya},
  journal={Advances in Neural Information Processing Systems},
  volume={35},
  pages={18442--18455},
  year={2022}
}

@inproceedings{li2022voxel,
  title={Voxel field fusion for 3d object detection},
  author={Li, Yanwei and Qi, Xiaojuan and Chen, Yukang and Wang, Liwei and Li, Zeming and Sun, Jian and Jia, Jiaya},
  booktitle={Proceedings of the IEEE/CVF Conference on Computer Vision and Pattern Recognition},
  pages={1120--1129},
  year={2022}
}

@article{liang2022bevfusion,
  title={Bevfusion: A simple and robust lidar-camera fusion framework},
  author={Liang, Tingting and Xie, Hongwei and Yu, Kaicheng and Xia, Zhongyu and Lin, Zhiwei and Wang, Yongtao and Tang, Tao and Wang, Bing and Tang, Zhi},
  journal={Advances in Neural Information Processing Systems},
  volume={35},
  pages={10421--10434},
  year={2022}
}

@inproceedings{chen2023focalformer3d,
  title={Focalformer3d: focusing on hard instance for 3d object detection},
  author={Chen, Yilun and Yu, Zhiding and Chen, Yukang and Lan, Shiyi and Anandkumar, Anima and Jia, Jiaya and Alvarez, Jose M},
  booktitle={Proceedings of the IEEE/CVF International Conference on Computer Vision},
  pages={8394--8405},
  year={2023}
}

@inproceedings{li2024gafusion,
  title={GAFusion: Adaptive Fusing LiDAR and Camera with Multiple Guidance for 3D Object Detection},
  author={Li, Xiaotian and Fan, Baojie and Tian, Jiandong and Fan, Huijie},
  booktitle={Proceedings of the IEEE/CVF Conference on Computer Vision and Pattern Recognition},
  pages={21209--21218},
  year={2024}
}

@inproceedings{yin2024fusion,
  title={Is-fusion: Instance-scene collaborative fusion for multimodal 3d object detection},
  author={Yin, Junbo and Shen, Jianbing and Chen, Runnan and Li, Wei and Yang, Ruigang and Frossard, Pascal and Wang, Wenguan},
  booktitle={Proceedings of the IEEE/CVF Conference on Computer Vision and Pattern Recognition},
  pages={14905--14915},
  year={2024}
}

@inproceedings{xie2023sparsefusion,
  title={Sparsefusion: Fusing multi-modal sparse representations for multi-sensor 3d object detection},
  author={Xie, Yichen and Xu, Chenfeng and Rakotosaona, Marie-Julie and Rim, Patrick and Tombari, Federico and Keutzer, Kurt and Tomizuka, Masayoshi and Zhan, Wei},
  booktitle={Proceedings of the IEEE/CVF International Conference on Computer Vision},
  pages={17591--17602},
  year={2023}
}

@inproceedings{jiao2023msmdfusion,
  title={Msmdfusion: Fusing lidar and camera at multiple scales with multi-depth seeds for 3d object detection},
  author={Jiao, Yang and Jie, Zequn and Chen, Shaoxiang and Chen, Jingjing and Ma, Lin and Jiang, Yu-Gang},
  booktitle={Proceedings of the IEEE/CVF conference on computer vision and pattern recognition},
  pages={21643--21652},
  year={2023}
}

@inproceedings{liu2023bevfusion,
  title={Bevfusion: Multi-task multi-sensor fusion with unified bird's-eye view representation},
  author={Liu, Zhijian and Tang, Haotian and Amini, Alexander and Yang, Xinyu and Mao, Huizi and Rus, Daniela L and Han, Song},
  booktitle={2023 IEEE international conference on robotics and automation (ICRA)},
  pages={2774--2781},
  year={2023},
  organization={IEEE}
}

@article{wang2023multi,
  title={Multi-sensor fusion technology for 3D object detection in autonomous driving: A review},
  author={Wang, Xuan and Li, Kaiqiang and Chehri, Abdellah},
  journal={IEEE Transactions on Intelligent Transportation Systems},
  year={2023},
  publisher={IEEE}
}

@article{cai2023bevfusion4d,
  title={BEVFusion4D: Learning LiDAR-Camera Fusion Under Bird's-Eye-View via Cross-Modality Guidance and Temporal Aggregation},
  author={Cai, Hongxiang and Zhang, Zeyuan and Zhou, Zhenyu and Li, Ziyin and Ding, Wenbo and Zhao, Jiuhua},
  journal={arXiv preprint arXiv:2303.17099},
  year={2023}
}

@article{li2024cofix3d,
  title={Co-Fix3D: Enhancing 3D Object Detection with Collaborative Refinement},
  author={Wenxuan Li and Qin Zou and Chi Chen and Bo Du and Long Chen},
  journal={https://arxiv.org/abs/2408.07999},
  year={2024}
}

@inproceedings{xu2021fusionpainting,
  title={Fusionpainting: Multimodal fusion with adaptive attention for 3d object detection},
  author={Xu, Shaoqing and Zhou, Dingfu and Fang, Jin and Yin, Junbo and Bin, Zhou and Zhang, Liangjun},
  booktitle={2021 IEEE International Intelligent Transportation Systems Conference (ITSC)},
  pages={3047--3054},
  year={2021},
  organization={IEEE}
}

@article{song2024robustness,
  title={Robustness-aware 3d object detection in autonomous driving: A review and outlook},
  author={Song, Ziying and Liu, Lin and Jia, Feiyang and Luo, Yadan and Jia, Caiyan and Zhang, Guoxin and Yang, Lei and Wang, Li},
  journal={IEEE Transactions on Intelligent Transportation Systems},
  year={2024},
  publisher={IEEE}
}

@inproceedings{song2025graphbev,
  title={Graphbev: Towards robust bev feature alignment for multi-modal 3d object detection},
  author={Song, Ziying and Yang, Lei and Xu, Shaoqing and Liu, Lin and Xu, Dongyang and Jia, Caiyan and Jia, Feiyang and Wang, Li},
  booktitle={European Conference on Computer Vision},
  pages={347--366},
  year={2025},
  organization={Springer}
}

@inproceedings{zeng2022lift,
  title={Lift: Learning 4d lidar image fusion transformer for 3d object detection},
  author={Zeng, Yihan and Zhang, Da and Wang, Chunwei and Miao, Zhenwei and Liu, Ting and Zhan, Xin and Hao, Dayang and Ma, Chao},
  booktitle={Proceedings of the IEEE/CVF conference on computer vision and pattern recognition},
  pages={17172--17181},
  year={2022}
}

@article{mu2025stereodetr,
  title={StereoDETR: Stereo-based Transformer for 3D Object Detection},
  author={Mu, Shiyi and Gu, Zichong and Ai, Zhiqi and Liu, Anqi and Gao, Yilin and Xu, Shugong},
  journal={IEEE Transactions on Circuits and Systems for Video Technology},
  year={2025},
  publisher={IEEE}
}

@article{jia2025dgfusion,
  title={DGFusion: Dual-guided Fusion for Robust Multi-Modal 3D Object Detection},
  author={Jia, Feiyang and Jia, Caiyan and Liu, Ailin and Xu, Shaoqing and Xia, Qiming and Liu, Lin and Yang, Lei and Gong, Yan and Song, Ziying},
  journal={IEEE Transactions on Circuits and Systems for Video Technology},
  year={2025},
  publisher={IEEE}
}

@article{feng2025iter3ddet,
  title={Iter3DDet: Depth-Guided Iterative Fusion and Refinement for Monocular 3D Object Detection},
  author={Feng, Cheng and Zhang, Congxuan and Chen, Zhen and Hu, Weiming and Lu, Ke and Ge, Liyue},
  journal={IEEE Transactions on Circuits and Systems for Video Technology},
  year={2025},
  publisher={IEEE}
}

@article{zhou2025fastpillars,
  title={Fastpillars: A deployment-friendly pillar-based 3d detector},
  author={Zhou, Sifan and Zhang, Xinyu and Chu, Xiangxiang and Zhang, Bo and Zhao, Ziyu and Lu, Xiaobo},
  journal={IEEE Transactions on Circuits and Systems for Video Technology},
  year={2025},
  publisher={IEEE}
}

@article{lin2025method,
  title={A method of time alignment in bev features for multimodal fusion object detection of intelligent vehicles},
  author={Lin, Chen and He, Zhicheng and Qiu, Yu and Huang, Yuanyi},
  journal={IEEE Transactions on Intelligent Transportation Systems},
  year={2025},
  publisher={IEEE}
}

@inproceedings{cai2023objectfusion,
  title={Objectfusion: Multi-modal 3d object detection with object-centric fusion},
  author={Cai, Qi and Pan, Yingwei and Yao, Ting and Ngo, Chong-Wah and Mei, Tao},
  booktitle={Proceedings of the IEEE/CVF international conference on computer vision},
  pages={18067--18076},
  year={2023}
}

@article{chen2024dsc3d,
  title={Dsc3d: Deformable sampling constraints in stereo 3d object detection for autonomous driving},
  author={Chen, Jiawei and Song, Qi and Guo, Wenzhong and Huang, Rui},
  journal={IEEE Transactions on Circuits and Systems for Video Technology},
  volume={35},
  number={3},
  pages={2794--2805},
  year={2024},
  publisher={IEEE}
}

@article{li2025tinyfusiondet,
  title={TinyFusionDet: Hardware-Efficient LiDAR-Camera Fusion Framework for 3D Object Detection at Edge},
  author={Li, Yishi and Zeng, Fanhong and Lai, Rui and Wu, Tong and Guan, Juntao and Zhu, Anfu and Zhu, Zhangming},
  journal={IEEE Transactions on Circuits and Systems for Video Technology},
  year={2025},
  publisher={IEEE}
}
}

\end{document}